\PassOptionsToPackage{protrusion=true,expansion=true}{microtype}
\PassOptionsToPackage{dvipsnames,table}{xcolor}
\documentclass[sigconf,nonacm]{acmart}
\usepackage[utf8]{inputenc}
\usepackage{microtype}
\usepackage{algorithm}
\usepackage[capitalize,nameinlink]{cleveref}
\newcolumntype{Y}{>{\raggedright\arraybackslash}X}
\usepackage[most]{tcolorbox}
\usepackage{newtxmath}
\makeatletter
\def\fps@algorithm{htbp}   
\makeatother
\usepackage{algpseudocode}
\usepackage{amsmath}
\usepackage{booktabs}
\usepackage{enumitem}
\usepackage{graphicx}
\usepackage{makecell}
\usepackage{colortbl}
\usepackage{multirow}
\usepackage{threeparttable}
\usepackage{pgfplots}
\pgfplotsset{compat=1.18}
\usepackage{overpic}
\usepackage{pifont}
\usepackage{array}
\newcolumntype{L}[1]{>{\raggedright\arraybackslash}p{#1}}

\usepackage{tabularx}

\AtBeginDocument{%
  }

\apptocmd{\thebibliography}{\sloppy}{}{}
\setcopyright{none}
\renewcommand\footnotetextcopyrightpermission[1]{}

\graphicspath{{figs/}} 
\begin{document}

\title[Perspective-Conditioned Spatial Reasoning from Omnidirectional Images]{Beyond Localization: A Comprehensive Benchmark of Perspective-Conditioned Spatial Reasoning in MLLMs from Omnidirectional Images}
\author{Yuangong Chen}
\orcid{0009-0008-0849-7652}
\affiliation{%
  \institution{The Hong Kong Polytechnic University}
  \city{Hong Kong}  
  \country{China}
}
\email{caleb.chen@connect.polyu.hk}

\author{Waikeung Wong}
\orcid{0000-0002-5214-7114}
\authornote{Corresponding author.}
\affiliation{%
  \institution{The Hong Kong Polytechnic University}
  \city{Hong Kong}
  \country{China}
}
\email{calvin.wong@polyu.edu.hk}

\author{Jiaxing Li}
\orcid{0000-0002-8572-2920}
\affiliation{%
  \institution{Guangzhou University}
  \city{Guangzhou}
  \country{China}
}
\email{jiaxing.li.cs@gmail.com}

\author{Ioannis Patras}
\orcid{0000-0003-3913-4738}
\affiliation{%
  \institution{Queen Mary, University of London}
  \city{London}
  \country{United Kingdom}
}
\email{i.patras@qmul.ac.uk}

\author{Xu Zheng}
\orcid{0000-0003-4008-8951}
\affiliation{%
\institution{Great Bay University}
\city{Dongguan}
\country{China}}
\email{zhengxu128@gmail.com}

\renewcommand{\shortauthors}{Yuangong Chen, Waikeung Wong, Jiaxing Li, Ioannis Patras, \& Xu Zheng.}

\begin{abstract}
Multimodal Large Language Models (MLLMs) show strong visual perception, yet remain limited in reasoning about space under changing viewpoints. We study this challenge as \textbf{P}erspective-\textbf{C}onditioned \textbf{S}patial \textbf{R}easoning (\textbf{PCSR}) in $360^\circ$ omnidirectional images, where broad scene coverage reduces ambiguity from partial observations without eliminating the need for viewpoint-dependent inference. To assess this capability, we introduce \textbf{PCSR-Bench}, a diagnostic benchmark of 84,373 QA pairs from 2,600 omnidirectional images across 26 indoor environments, 
organized into eight tasks under three cognitive groups--- 
\emph{Perception}, \emph{Spatial}, and
\emph{advanced PCSR}. 
We evaluate 14 representative MLLMs and observe a substantial perception--reasoning gap: accuracy reaches 57.59\% on Limited Field-of-View Reasoning (T7) but drops to 13.49\%, 7.13\%, and 0.64\% 
on Relative Direction (T2), Egocentric Rotation (T4), and open-ended Compositional Directional Chains (T3), respectively.
To probe the plasticity of this gap, we conduct an RL-based diagnostic study on a 7B-scale model. Reward shaping improves a matched 7B baseline from 31.10\% to 60.06\% under a controlled setting, suggesting that PCSR exhibits partial plasticity rather than being fully immutable. Still, these gains are task-selective, sensitive to reward design, and partially dependent on the evaluation protocol. These results position PCSR as a key bottleneck in current MLLMs and highlight meaningful yet bounded room for recovery under targeted optimization. Details and access are available at \url{https://github.com/Caleb-ychen/PCSR-Benchmark}.
\end{abstract}




\begin{teaserfigure}
  \centering  
  \includegraphics[width=0.85\textwidth]{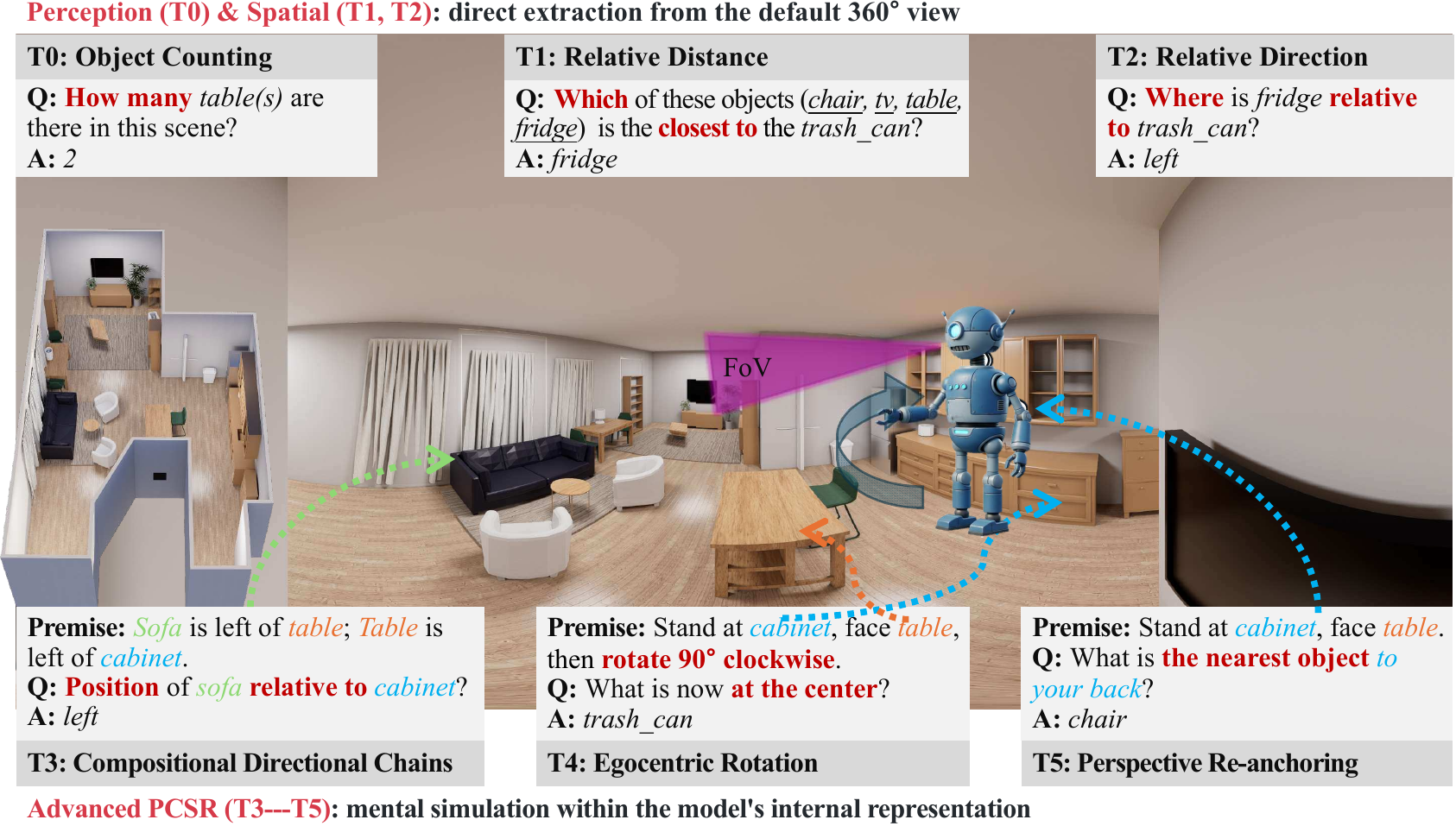}
  \caption{Diagnostic task structure of PCSR-Bench. Examples cover Object Counting (T0) from the Perception group, Relative Distance (T1) and Relative Direction (T2) from the Spatial group, and the three advanced PCSR tasks (T3--T5). The remaining Perception tasks, Egocentric Distortion (T6) and Limited Field-of-View Reasoning (T7), are not illustrated.}
  \Description{A systematic diagram of the PCSR-Bench framework. Three panels above the panorama show tasks
  answerable directly from the default view: Object Counting (T0), Relative
  Distance (T1) and Relative Direction (T2). Three panels below show the
  advanced perspective-conditioned tasks: Compositional Directional Chains (T3),
  Egocentric Rotation (T4) and Perspective Re-anchoring (T5).}
  \label{fig:tasks}
\end{teaserfigure}


\maketitle

\section{Introduction}

A seemingly simple request---Take the plate on the right side of the table to the kitchen---requires an agent not only to locate the table and the plate, but also to interpret the spatial relation \emph{right side} with respect to the speaker’s position and viewpoint. As the observer moves or turns, relations such as left/right, front/back, and even visibility may change accordingly. This reflects a fundamental property of spatial cognition: viewpoint change requires more than static object recognition, and has long been studied as a problem of mental transformation and internal spatial modeling {\cite{shepard1971mental, goral2024seeing, wang2026egocentric}}. Recent evidence further suggests that current vision-language models exhibit a persistent egocentric bias, tending to anchor on the primary camera viewpoint and showing limited robustness in Level-2 visual perspective taking \citep{gao2025vision, goral2026beyond, yan-etal-2025-survey}.

We term this capability \textbf{P}er\-spec\-tive-\textbf{C}on\-di\-tioned \textbf{S}pa\-tial \textbf{R}ea\-son\-ing (\textbf{PCSR}): updating, re-anchoring, and inferring spatial relations
under a changed or hypothesized observer position, orientation, and viewpoint.
Following prior navigation work \cite{Banino2018VectorBasedNavigation}, 
we use \emph{observer-conditioned spatial representation} to denote the internal
scene representation 
that supports such updates.
PCSR-Bench isolates this capability as a diagnostic
prerequisite for embodied settings rather than simulating of
full robotic control.

Evaluating PCSR therefore requires reasoning under viewpoint change rather than merely recognizing spatial relations from a fixed camera view. However, existing benchmarks---notably including even recent omnidirectional frameworks \cite{Dongfang_2026_CVPR, egonight}---do not directly target the core challenges of PCSR. To bridge this gap, we introduce PCSR-Bench, a benchmark designed to test whether models can recompute spatial relations under changed observer conditions. A central design goal of PCSR-Bench is to reduce the confound between incomplete observation and genuine reasoning failure.

PCSR-Bench is built on 360\textdegree{} panoramic images and 3D scene annotations,
giving models near-complete scene coverage
from a single observation and thereby reducing ambiguity caused by partial observation,
while still requiring viewpoint-dependent inference \cite{dong2023panocontext}. This design helps separate failures of observation from reasoning. The benchmark is organized into three cognitive groups
of increasing demand: 
Perception (\textbf{T0}, \textbf{T6} and \textbf{T7}), 
Spatial (\textbf{T1}---\textbf{T2}), and
advanced PCSR \textbf{T3}---\textbf{T5}. 
In this way, PCSR-Bench evaluates not only target localization but also whether models form dynamic, observer-conditioned spatial representations.

Systematic zero-shot evaluation on PCSR-Bench reveals a clear gap in current MLLMs’ spatial abilities. 
Models perform nontrivially on foundational perception tasks, but degrade sharply on advanced PCSR tasks. This pattern recurs across advanced PCSR subtasks, indicating a shared systematic failure across tasks. 

To probe the plasticity of this gap, we explore a reinforcement-learning-based intervention on PCSR-Bench. Recent work shows that RL can elicit stronger reasoning behavior in both language and multimodal models \citep{guo2025deepseek, huang2026vision}. Following this line, we use RL not to claim that the problem is solved, but to diagnose whether targeted training signals can partially recover it. 
Our results show that suitable reward design and task feedback improve performance on some advanced PCSR, but gains remain uneven and the broader gap between foundational perception and advanced PCSR persists. This suggests that perspective-conditioned spatial reasoning is not an automatic byproduct of general visual capacity, it requires explicit modeling, targeted training, and dedicated evaluation.

In summary, we make three contributions. First, we \textbf{formally define PCSR} as the ability to update, re-anchor, and infer spatial relations under changing viewpoints, establishing it as a distinct core capability beyond fixed views.
Second, we \textbf{introduce PCSR-Bench}, a 3D-grounded diagnostic benchmark built on omnidirectional observations and a structured task suite spanning foundational perception to advanced perspective-conditioned reasoning.
Third, we use PCSR-Bench to provide a \textbf{diagnosis of current MLLMs} and to study the \textbf{partial plasticity of PCSR under RL-based intervention}, revealing both a stable perception--reasoning gap and limited but non-negligible recoverability.

\section{Related Work}
\label{gen_inst}
\subsection{Visual Spatial Reasoning Benchmarks}
\label{sec:rw_foundations_and_static}
Prior work has developed a progression of benchmarks to evaluate spatial reasoning in MLLMs \cite{lee2025perspective}, ranging from controlled synthetic datasets to more explicit spatial diagnostics. 
Early datasets such as CLEVR \cite{johnson2017clevr}, ShapeWorld \cite{kuhnle2017shapeworld}, and NLVR \cite{suhr2017nlvr} emphasized controlled compositional reasoning in synthetic settings, whereas NLVR2 \cite{suhr2019nlvr2} and GQA \cite{hudson2019gqa} moved to natural images with richer semantics and scene structure \citep{suhr2019nlvr2,hudson2019gqa}. 
Later work made spatial reasoning more central: VSR focuses on fine-grained spatial-language understanding \cite{liu2023vsr}, SpatialVLM \cite{chen2024spatialvlm} and TopViewRS \cite{li2024topviewrs} study spatial VQA and map reasoning, and OSR-Bench \cite{Dongfang_2026_CVPR}, CoSpace \cite{zhu2025cospace}, and Spatial457 \citep{wang2025spatial457} extend evaluation to omnidirectional, continuous-view, and 6D spatial settings.

These benchmarks substantially improve the evaluation of foundational spatial perception, including counting, relative distance estimation, directional judgment, and object-centric spatial reasoning \cite{hua2026unleashing}. However, most remain tied to a \textit{fixed observer perspective}. Even with panoramic or top-down inputs, models typically reason from the provided viewpoint rather than re-anchoring the observer under a hypothetical one \cite{geirhos2020shortcut, goyal2017making}. Thus, existing benchmarks are effective for measuring static spatial perception, but less suited for isolating perspective-conditioned spatial reasoning.
A side-by-side comparison with prior benchmarks is given in
Appendix~E (Table~E.1). 
PCSR-Bench adopts a panorama-disjoint split shared
with OSR-Bench and 360-R1; cross-scene generalization is reported separately by VSI-Bench / MMSI-Bench (See Appendix A.7).

\subsection{Perspective-Conditioned Reasoning} \label{sec:rw_perspective}
Recent work in embodied AI and multimodal reasoning has begun to touch on perspective-conditioned spatial reasoning (PCSR), grounded on classic notions of mental rotation and viewpoint transformation \citep{shepard1971mental,das2017embodied,yang2025thinking}. Yet in most current benchmarks, PCSR is entangled with partial observability, sequential perception, and navigation or action planning. 
This confounding complexity makes it difficult to isolate failures specifically from viewpoint re-anchoring itself rather than from related factors such as missing observations or unstable scene representations \cite{zhu2026video}.
While prior work on omnidirectional scene understanding has demonstrated that wide-field observations can mitigate blind spots \cite{zhang2014panocontext, da2023omnidirectional, OmniSAM}, such methods do not by themselves provide a principled diagnosis of a model's ability to recompute spatial relations under shifting observer conditions. We isolate PCSR in a controlled setting, enabling a direct assessment of perspective-conditioned reasoning.

\subsection{Enhancing Spatial Reasoning in MLLMs}
\label{sec:RL_visionSR}
While supervised fine-tuning has been the dominant paradigm for aligning MLLMs, it often favors surface-level pattern matching over robust multi-step inference \citep{burns2023weak, ouyang2022training}.
Recently, Reinforcement Learning (RL) has shown strong potential for
eliciting deliberative reasoning in both language and multimodal models \citep{zheng2025multimodalspatialreasoninglarge}. By optimizing outcome- and process-level rewards \citep{sun2024aligning, yu2024rlhfv}, RL encourages extended chains of thought (CoT),  yielding substantial gains in mathematical, coding, and visual reasoning tasks \cite{wei2022chain, lightman2023let}.

The extent to which RL can similarly improve perspective--conditioned spatial reasoning remains unclear. Unlike general visual QA, PCSR requires observer-conditioned spatial representation,
performing coordinate transformations,
and re-anchoring spatial relation under changed viewpoint. These mirror the core challenges of object-state tracking  and spatial updating in embodied settings \citep{szot2021habitat}.
Recent benchmarks further suggest that current MLLMs struggle consistently when required to reason from non-camera or alternative viewpoints, exhibiting systematic egocentric bias and degraded performance across first-person and cross-view settings \citep{cheng2024egothink, li2025viewspatial, wang2026egocentric}.

In this context, RL is not only a method of improving performance but also a diagnostic tool for assessing model plasticity. Thus, we employ reward-guided RL as a diagnostic intervention to test the recoverability of PCSR
under controlled evaluation settings.

\section{PCSR-Bench Construction}
\label{PCSR-Bench}

\begin{figure*}[!htbp]
    \centering
    \begin{overpic}[width=0.9\textwidth]{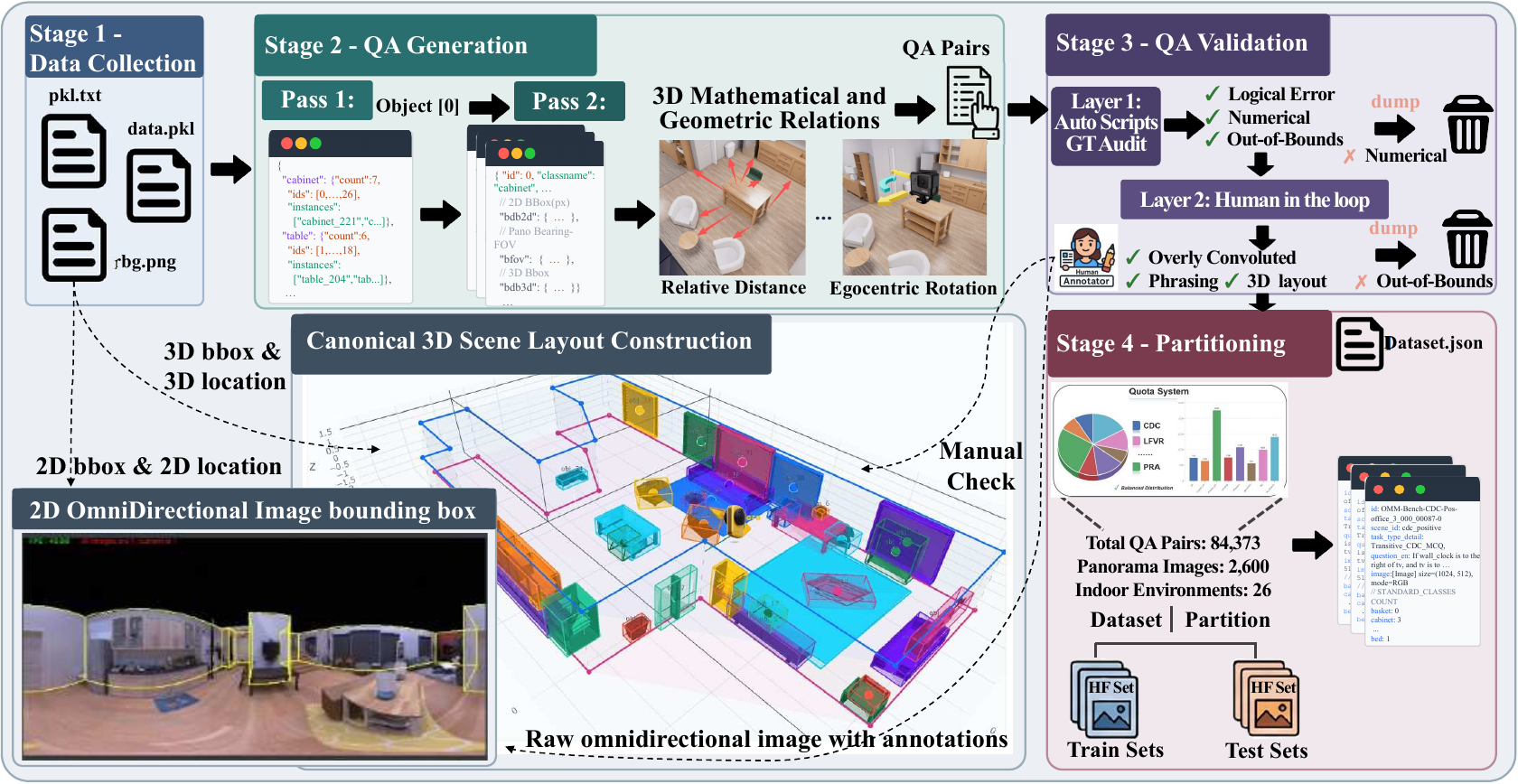} 
    \end{overpic}
    \caption{PCSR-Bench four-stage construction pipeline (App.~A): \ding{192} Stage 1 -- Data Collection:
    2{,}600 panoramas across indoor scenes paired with 3D meshes and per-object annotations.
    \ding{193} Stage 2 -- QA Generation: two-pass template-driven QA construction across eight tasks T0--T7.
    \ding{194} Stage 3 -- QA Validation: two-layer checks (automated + human).
    \ding{195} Stage 4 -- Partitioning:
    panorama-disjoint split into train / test items.}
    \Description{Four-stage horizontal pipeline: panorama + 3D-mesh data
    collection, two-pass QA generation over eight tasks, two-layer
    automatic-and-human QA validation, and panorama-disjoint train/test
    partitioning released as a public dataset.}
    \label{fig:omm_bench}
\end{figure*}

PCSR-Bench combines a scalable construction pipeline, a structured task suite, and a rigorous evaluation protocol 
(Figure~\ref{fig:omm_bench}).

\subsection{Task Suite}
\label{subsec:tasktaxonomy}
PCSR-Bench is organized into eight diagnostic tasks (T0---T7) across three groups. As illustrated in Figure~\ref{fig:tasks}, it uses 2D/3D scene annotations to generate task instances, while evaluation is performed solely on raw panoramic inputs. 

\textbf{Perception tasks} (T0, T6, T7) evaluate baseline visual grounding and robustness to omnidirectional imaging. T0 (\emph{Object Counting},Count) tests object identification and counting. 
T6 (\emph{Egocentric Distortion}, Ego) tests reasoning under omnidirectional projection distortion.
T7 (\emph{Limited Field-of-View Reasoning}, LFVR) assesses occlusion and visibility judgments under restricted field-of-view (FoV) constraints.

\textbf{Spatial tasks} (T1, T2) evaluate object-to-object relations read directly
from the $360^\circ$ observation. T1 (\emph{Relative Distance}, Dist) evaluates relative
distance judgments between objects. T2 (\emph{Relative Direction}, Direction) evaluates
relative directional judgments between objects.

\begin{sloppypar}
\textbf{Advanced PCSR tasks} (T3--T5) probe whether models can update, re-anchor, and constrain spatial judgments under changed observer conditions (See Figure~\ref{fig:tasks}).
T3 (\emph{Compositional Directional Chains}, CDC) tests multi-step relational composition, including transitive and compositional inference.
T4 (\emph{Egocentric Rotation}, ERR) tests spatial updating under observer rotation.
T5 (\emph{Perspective Re-anchoring}, PRA) tests whether models can re-anchor spatial judgments to a new observer viewpoint. 
\end{sloppypar}
A failure-mode coverage matrix (Table~E.2) and verbatim worked examples (Appendix~E.2) are provided.

\subsection{Benchmark Construction}
\label{sec:benchmark-construction}
\textbf{Data Collection and Canonical 3D Scene Layout Construction}. 
We build PCSR-Bench on \textbf{ReplicaPano} \citep{dong2023panocontext}, a photorealistic indoor $360^\circ$ panoramic dataset derived from Replica  \citep{straub2019replica}, whose 26 environments span
apartments, offices, standalone rooms, and a hotel room (see supplementary scene-diversity examples).
Because panoramic
projection distorts planar geometry, we derive spatial relations
from 3D annotations rather than 2D appearance alone, supporting
reliable QA generation and verification \cite{coors2018spherenet, tateno2018distortion}.

We transform the raw metadata of each panorama into a canonical 3D scene layout that unifies room geometry, camera pose, object instances, geometric extents, semantic categories, and instance identities in a single operable space. This layout enables exact geometric computation, rendered 3D inspection, and downstream QA generation.

\textbf{3D Geometry-Constrained QA Generation}. Unlike prior efforts \cite{chen2024spatialvlm} that attempt to lift Internet-scale 2D images into approximate 3D representations using expert vision models, our QA generation is strictly governed by explicit 3D mathematical and geometric relations (e.g., Euclidean distances, vector dot products, and rotation matrices) derived from the canonical layout. This design forces models to perform true spatial reasoning rather than relying on image-plane heuristics. 

Specifically, for tasks that admit both response formats (T1, T3, T5), we generate paired evaluation items in two formats: \ding{192} Open-ended questions that test absolute localization and zero-shot reasoning without candidate hints, and \ding{193} multiple-choice questions (MCQ) with reasoning-critical distractors. 
Instead of sampling random negative options, we computationally design hard distractors based on exact 3D spatial relations. For instance, in relative distance (T1), the distractor is explicitly set as the second-closest object; and in compositional directional chains (T3), distractors are objects that satisfy only partial logical hops. This forces the model to execute precise numerical comparisons and complete logical deductions rather than rely on simple elimination. The implementation details of QA pair generation can be found in Appendix A.

\textbf{Quality Assurance and Validation}. 
The generated data undergoes a two-stage protocol. First, an \textbf{automatic audit} 
checkes schema consistency, logical validity, 
answer-format compliance, and QA---panoramic alignment against the canonical 3D source. Second, \textbf{task-stratified human validation} assesses naturalness,
ambiguity, and visual answerability from panoramas alone, ruling out items solvable by superficial 2.5D cues. 
A stratified audit of 850 QA items across 17 sub-tasks yielded
100.00\% naturalness, 5.65\% ambiguity, 0.35\% reformulation, and
5.41\% removal candidates, with Cohen's $\kappa{=}0.682$ (substantial). Full protocol: Appendix~A.

\begin{figure}[!htbp]
    \centering
    \includegraphics[width=0.80\columnwidth]{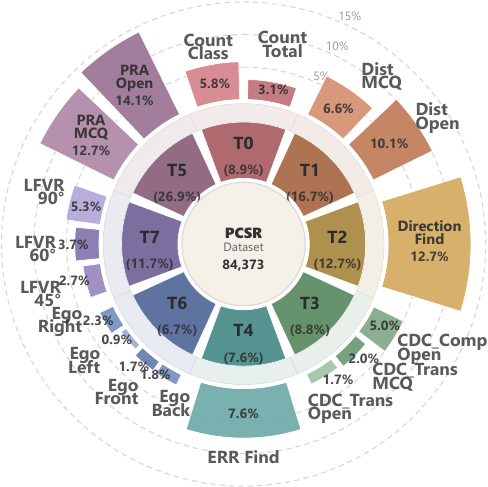}
    \caption{Task distribution of PCSR-Bench, organized into Perception (T0, T6, T7), Spatial (T1, T2), and advanced
    PCSR (T3--T5).
    The inner ring shows task proportions, the outer rings break each task down by question format and, for advanced subtask type, by transitive vs.\ compositional subtype. 
    Acronym are expanded directly on the chart.}
    \Description{A multi-level rose chart summarizing PCSR-Bench's 84,373
    question-answer pairs across eight tasks. By share of the dataset,
    the Perception group (T0, T6, T7) accounts for 27.3 percent, the Spatial group
    (T1, T2) for 29.4 percent, and the advanced PCSR group (T3 to T5) for 43.3
    percent. The largest individual tasks are T5 Perspective Re-anchoring at 26.9
    percent and T1 Relative Distance at 16.7 percent. The inner ring shows task
    proportions; outer rings break each task down by question format and, for
    advanced tasks, by subtype.
    }    
    \label{fig:pcsr_distribution}
\end{figure}

\textbf{Dataset Partitioning}. PCSR-Bench comprises 84,373 QA pairs built from 2{,}600 panoramas in
26 indoor scenes 
(Figure~\ref{fig:pcsr_distribution}), split at the
panorama level so that
no image contributes
to both sides. 
Zero-shot evaluation (\S\ref{sub:zero_shot}) uses the shared test split for
all
14 MLLMs; the RL pilot (\S\ref{sec:rl}) uses a 
$\sim$10\%
task-stratified subsample of the training split
for single-epoch GRPO. Per-task and per-format breakdowns: Appendix~A.

\section{Zero-Shot Evaluation on PCSR-Bench}
\label{sub:zero_shot}

\subsection{Models and Setup}
\label{subsec:model_selection}
We evaluated 14 MLLMs across three capability tiers.
\ding{192}~\textbf{Tier-1} ($\geq$72B/API): InternVL2.5-78B-Instruct, Qwen2.5-VL-72B, Qwen-VL-Max;
\ding{193}~\textbf{Tier-2} (10B--40B): InternVL3.5-38B, Qwen3-VL-32B-Instruct, and Llama-3.2-11B-Vision-Instruct;
\ding{194}~\textbf{Tier-3} ($<$10B): InternVL2.5-8B, InternVL3.5-8B, Qwen2.5-VL-7B-Instruct, Qwen3-VL-8B, MiniCPM-V-2.5, LLaVA-v1.6-7B, LLaVA-OneVision-7B, Janus-Pro-7B.
Compass (Ours) trains three GRPO variants on Qwen2.5-VL-7B-Instruct (see \S\ref{sec:rl}).

All models are evaluated on the PCSR-Bench test split (\S\ref{sec:benchmark-construction}).
For cross-model comparability, we use a unified zero-shot prompt (shown below) with deterministic hard-greedy top-1 decoding. Architecture-compatible implementations are used only where needed for stable generation, without changing the evaluation policy; Details are deferred to Appendix~B.
\begin{center}
\begin{tcolorbox}[
    width=0.86\columnwidth,
    colback=gray!5,
    colframe=gray!20,
    arc=1pt,
    boxrule=0.5pt,
    left=5pt, right=5pt, top=5pt, bottom=5pt,
    halign=flush left
]
\small
\textit{\{Question\}} \par
Output the thinking process in \texttt{<think>} and \texttt{</think>},
and the final answer (a single word or number) in
\texttt{<answer>} and \texttt{</answer>} tags.
\end{tcolorbox}
\end{center}
For models with native CoT capability (e.g., InternVL3.5, Qwen3-VL), this template enables structured reasoning without additional system prompts. All 14 MLLMs share identical visual inputs, prompt templates,
and inference infrastructure (Table B.1);
differences are limited to the model weights.
\begin{center}
\begin{tcolorbox}[
    width=0.94\columnwidth,
    colback=gray!5, colframe=gray!20,
    arc=1pt, boxrule=0.5pt,
    left=5pt, right=5pt, top=4pt, bottom=4pt,
    halign=flush left
]
\small
\textbf{Protocol Configuration (unified across all 14 MLLMs).}
\emph{Backend:} lmdeploy / vLLM depending on architecture compatibility;
decoding fixed to hard-greedy top-1.
\emph{Visual budget:} \texttt{MAX\_PIXELS}=602{,}112 (native) or 652{,}680
(fallback when tiling requires alignment).
\emph{Generation budget:} 512 tokens for MCQ, 1024 for open-ended.
\emph{Prompt template:} shown above; no per-model system prompts.
\emph{Answer parser precedence:} \texttt{<answer>...</answer>} $\to$
first 200 characters $\to$ full trimmed output.
\emph{Scoring:} strict vs.\ relaxed (see \S\ref{sec:exp_setup}).
Full per-model instantiations: Table~B.1
(Appendix~B.1).
\end{tcolorbox}
\end{center}

\subsection{Scoring Protocol}
\label{sec:exp_setup}
Exact-string matching under-reports performance in zero-shot multimodal evaluation, since semantically correct answers vary in formatting, wording, and instance naming under verbose CoT outputs~\cite{antol2015vqa,liu2024mmbench}.
We therefore adopt a \emph{Semantic-Aligned Evaluation Scorer}
$\mathcal{M}$ and define task-level accuracy as Eq.~(\ref{eq:acc}), with a strict or relaxed variant depending on the reported metric.
\begin{equation}
  \mathrm{Acc}
  = \frac{1}{N}\sum_{i=1}^{N}
    \mathbb{I}\!\left[
      \mathcal{M}\!\left(
        a_{\mathrm{pred}}^{(i)},\;a_{\mathrm{gt}}^{(i)}
      \right) = \mathrm{True}
    \right]
  \label{eq:acc}
\end{equation}
Before scoring, outputs are standardized by a lightweight extraction step: We strip the \texttt{<think>} content, 
parse the \texttt{<answer>} block, and fall back to the first 200 characters when no answer tag is present. Strict accuracy requires exact match after basic normalization; relaxed accuracy additionally applies ordered surface-form heuristics (intent canonicalization, synonym expansion, instance-to-class relaxation, bounded substring matching). Relaxed targets surface-form noise, not semantic over-acceptance: no paraphrase, no partial credit. Full rules and an end-to-end walkthrough: Appendix~B.2.

\subsection{Main Results and Analysis}
\label{sec:main_results}

Table~\ref{tab:zero_shot_relaxed} shows zero-shot top-1 relaxed accuracy on PCSR-Bench for 14 representative MLLMs in eight spatial-reasoning tasks.
Models are grouped into three tiers by parameter scale or API access.

\begin{table}[ht]
\centering
\caption{%
  \textbf{Zero-shot top-1 relaxed accuracy (\%) on PCSR-Bench.}
  Results are reported for 14 MLLMs across eight task categories grouped by cognitive type.  
  \textbf{Bold} = best in column; \underline{underline} = second-best.
  $^\dagger$: proprietary API model.
  $^\ddagger$: evaluated with CoT prompting.
}
\label{tab:zero_shot_relaxed}
\scriptsize 
\setlength{\tabcolsep}{1pt} 
\renewcommand{\arraystretch}{1.1} 
\begin{tabularx}{\columnwidth}{l @{\extracolsep{\fill}} ccc cc ccc | c}
\toprule
\noalign{\vskip -2pt}
\multirow{2}{*}{\textbf{Model}}
& \multicolumn{3}{c}{\textbf{Perception}}
& \multicolumn{2}{c}{\textbf{Spatial}}

& \multicolumn{3}{c|}{\textbf{Advanced PCSR}}
& \multirow{2}{*}{\textbf{Avg}} \\

\noalign{\vskip -2pt}
\cmidrule(lr){2-4} \cmidrule(lr){5-6} \cmidrule(lr){7-9}
\noalign{\vskip -2pt}
& \textbf{T0} & \textbf{T6} & \textbf{T7}
& \textbf{T1} & \textbf{T2}
& \textbf{T3} & \textbf{T4} & \textbf{T5}
& \\
\noalign{\vskip -2pt}


\midrule
\rowcolor{gray!15}
\multicolumn{10}{l}{\textbf{Tier-1:}~~$\geq$70\,B parameters or proprietary API} \\
\midrule
InternVL2.5-78B
  & \underline{44.06} & 29.45          & \textbf{83.74}
  & 26.45             & 15.49
  & 54.15 & 9.42           & 21.99
  & \textbf{35.59} \\
Qwen-VL-Max$^\dagger$
  & 39.91             & \underline{38.94} & 65.85
  & \underline{29.05} & \underline{17.03}
  & \underline{54.74}             & 15.07 & 21.03
  & \underline{35.20} \\
Qwen2.5-VL-72B-Instruct
  & 36.45             & 34.11          & 61.34
  & 26.21             & \textbf{17.36}
  & \textbf{56.96}    & \textbf{17.50} & 22.88
  & 34.10 \\
\midrule
\rowcolor{gray!15}
\multicolumn{10}{l}{\textbf{Tier-2:}~~10--40\,B parameters} \\
\midrule
Qwen3-VL-32B-Instruct$^\ddagger$
  & 34.72             & 38.44          & 58.45
  & 28.93             & 15.82
  & 31.93             & \underline{17.09} & 20.49
  & 30.73 \\
InternVL3.5-38B
  & 34.37             & 26.12          & 63.50
  & 28.64             & 15.09
  & 35.20             & 2.42           & 22.53
  & 28.49 \\
Llama-3.2-11B-Vision-Instruct$^\ddagger$
  & 32.64             & 34.11          & 59.17
  & 16.98             & 13.63
  & 35.20             & 3.90           & 20.26
  & 26.99 \\
\midrule
\rowcolor{gray!15}
\multicolumn{10}{l}{\textbf{Tier-3:}~~$<$10\,B parameters} \\
\midrule
Qwen3-VL-8B-Instruct
  & 38.29             & \textbf{40.93} & 46.52
  & \textbf{30.47}    & 16.46
  & 51.58             & 4.44           & 17.95
  & 30.83 \\
MiniCPM-Llama3-V-2.5
  & \textbf{62.05}    & 32.45          & 36.86
  & 20.18             & 6.24
  & 42.22             & 12.79         & 22.03
  & 29.35 \\
InternVL2.5-8B
  & 33.68             & 25.96          & \underline{75.70}
  & 15.98             & 13.30
  & 45.26             & 4.04           & 20.18
  & 29.26 \\
Qwen2.5-VL-7B-Instruct
  & 37.02             & 23.29          & 63.78
  & 19.23             & 13.38
  & 44.44             & 3.77           & 20.07
  & 28.12 \\
deepseek-ai/Janus-Pro-7B
  & 27.22             & 31.11          & 63.32
  & 23.79             & 12.81
  & 28.42             & 3.23           & \underline{23.30}
  & 26.65 \\
InternVL3.5-8B
  & 25.72             & 16.64          & 54.20
  & 20.77             & 13.06
  & 49.01             & 2.02           & 15.65
  & 24.63 \\
llava-v1.6-vicuna-7b-hf$^\ddagger$
  & 26.99             & 23.79          & 64.95
  & 18.76             & 5.76
  & 26.32             & 1.08           & 19.11
  & 23.34 \\
llava-onevision-qwen2-7b-ov-hf
  & 20.76             & 29.12          & 8.85
  & 20.53             & 13.38
  & 46.20             & 3.10           & \textbf{26.34}
  & 21.03 \\
\midrule

\rowcolor{gray!15}
\multicolumn{10}{l}{\textbf{Reference baselines \& self-auditing statistics}} \\
\midrule
\textbf{Random (per sub-task)}
  & 21.66 & 50.00 & 50.00
  & 4.96 & 12.50
  & 25.00 & 6.67  & 18.48
  & 23.66 \\
\textbf{95\% CI (1{,}000 boot.)$^\dagger$}
  & $\pm$1.85 & $\pm$2.31 & $\pm$1.16
  & $\pm$1.23 & $\pm$1.51
  & $\pm$2.73 & $\pm$0.87 & $\pm$0.85
  & $\pm$0.58 \\
\midrule
\textbf{Overall (all records)}
  & 35.28 & 30.32 & 57.59
  & 23.28 & 13.49
  & 42.97 & 7.13  & 20.99
  & 28.88 \\
\bottomrule

\end{tabularx}
\begin{minipage}{\linewidth}
\footnotesize
\raggedright
\textit{Note.}
T0--T7 denote Object Counting, Relative Distance, Relative Direction,
Compositional Directional Chains (CDC), Egocentric Rotation (ERR),
Perspective Re-anchoring (PRA), Egocentric Distortion (Ego),
and Limited Field-of-View (LFVR), respectively.
\textbf{Avg} column: arithmetic mean of the 8 task-level accuracies
in the same row (Macro-Task per model).
\textbf{Overall (all records):} per-task pooled accuracy across all
14 models; since the test set is shared, this equals the equal-model-weight
average (Micro = Macro-Model). Its Avg cell (28.88\%) equals the mean
of the 8 preceding cells.
\textbf{Random (per sub-task):} chance-level accuracy per sub-task, computed as 1/K where K is the effective answer-space size (MCQ option count for MCQ sub-tasks, empirical unique-GT cardinality for open sub-tasks).
\textbf{95\% CI:} percentile bootstrap with $B{=}1{,}000$ resamples over
all test items (seed 42)~\cite{efron1993introduction}.
\end{minipage}
\end{table}

\noindent \textbf{Scale helps, but architecture and generation matter more.}
While larger models tend to perform better, we observe a clear decoupling between scale and performance on PCSR-Bench.
Among Tier-1 models ($\geq$70\,B or proprietary), InternVL2.5-78B achieves the highest average (35.59\%), followed by Qwen-VL-Max (35.20\%) and Qwen2.5-VL-72B (34.10\%).
Notably, the strongest Tier-3 model, Qwen3-VL-8B (30.83\%), surpasses all Tier-2 models despite having 4--5$\times$ fewer parameters, outperforming Qwen3-VL-32B (30.73\%) and InternVL3.5-38B (28.49\%).
Similarly, within Tier-3, MiniCPM-V-2.5 (29.35\%) and InternVL2.5-8B (29.26\%) exceed all Tier-2 models except Qwen3-VL-32B.
These cross-tier inversions indicate that improvements in model design and training can offset, or even surpass, gains from scale, making parameter count a weak predictor of PCSR performance.

\noindent \textbf{A persistent perception--reasoning gap.}
Perception tasks are substantially easier than advanced or compositional reasoning tasks across all models.
T7 (LFVR) achieves the highest pooled accuracy (57.59\%), with InternVL2.5-78B reaching 83.74\%.
In contrast, T4 (ERR) is the most challenging (7.13\%; best: 17.50\%), followed by T2 (Direction) at 13.49\% (best: 17.36\%).
T3 (CDC) requires careful interpretation due to strong internal disparity: the open-ended compositional subset achieves only 0.64\% relaxed accuracy, whereas
the transitive subsets are much higher;
see Appendix~C.1 for the full breakdown.
Accordingly, the pooled T3 score (42.97\%) masks a sharp split between near-zero compositional open-ended performance and much stronger transitive inference.
These results indicate that current VLMs handle transitive chain inference far better than compositional spatial reasoning in open-ended settings.

\noindent \textbf{Instruction-following inconsistencies distort per-task profiles.}
Two models exhibit anomalous per-task patterns that reflect output-format sensitivity rather than genuine capability differences.
LLaVA-OneVision-7B, with an overall average of only 21.03\%,
scores 8.85\% on T7~(LFVR)---48.7\,pp below the overall pooled T7 accuracy 57.59\% ---yet
simultaneously records the best T5~(PRA) score across all 14 models (26.34\%),
indicating selective rather than uniform format failures.
Conversely, MiniCPM-V-2.5 achieves the best T0~(Counting) score of any model (62.05\%, roughly 27\,pp above the T0 overall of 35.28\%) while collapsing to 6.24\% on T2~(Direction), the second-lowest value across all models.
These cross-task disparities suggest that per-task performance is shaped not only by underlying reasoning ability, but also by task-specific factors that are not fully captured by a single task score. This observation motivates the robustness analysis in the next paragraph.

\noindent \textbf{Sensitivity to scoring protocol.}
\label{sec:robustness}
To further probe the reliability of these measurements, we compare strict and relaxed accuracy across all models (see Appendix~C for the full comparison). Relaxed scoring consistently yields higher accuracy than strict scoring across all 14 models. Averaged at the model level, strict accuracy is 24.08\%, whereas relaxed accuracy rises to 28.88\%, corresponding to a mean gain of 4.80 pp. The largest increase is observed for LLaVA-OneVision-7B (+20.59 pp), suggesting that exact-match evaluation can substantially undercount semantically correct predictions for some systems. At the same time, the gains for several other models remain comparatively modest, indicating that sensitivity to the scoring rule is uneven rather than universal.

While these zero-shot results highlight systematic failures, the following section explores the partial plasticity of these gaps through a diagnostic reinforcement learning pilot study.

\section{A Diagnostic RL Pilot on a 7B Backbone}
\label{sec:rl}

To probe whether the observed reasoning gap is at least partially plastic, we conduct a reinforcement-learning-based intervention by fine-tuning Qwen2.5-VL-7B-Instruct  on the PCSR-Bench training split using GRPO
\cite{shao2024deepseekmath}. This section serves as a diagnostic pilot study rather than a full solution to PCSR. Instead of numerically replacing the zero-shot results in \S4, it examines whether targeted reward design can improve perspective-conditioned spatial reasoning under a unified evaluation stack.
Unless otherwise noted, all within-section comparisons are made under this controlled stack. Detailed training hyperparameters are deferred to Appendix~D.

\subsection{Training Setup}
\label{subsec:rl_setup}
\noindent\textbf{Base model:}
We use Qwen2.5-VL-7B-Instruct as the starting checkpoint.
\textbf{Training data:}
Training uses the PCSR-Bench training split, comprising 7,499 QA pairs spanning all eight task types (T0--T7) with a balanced mix of multiple-choice and open-ended questions.
\textbf{Image processing:}
Images are processed through the Qwen2.5-VL dynamic-resolution pipeline with $\texttt{max\_pixels}=652{,}680$. Since the native panoramic resolution is $1024\times512$ (524,288 pixels), no downscaling is required; each spatial dimension is rounded to a multiple of 28, yielding an effective processing resolution of $1036\times504$.
\textbf{Training configuration:}
All variants are trained for one epoch with maximum completion length 512 on 8$\times$A800 (80GB) GPUs using DeepSpeed ZeRO Stage~2, FlashAttention-2, and bfloat16 precision. The per-device batch size is 1 with gradient accumulation over 4 steps (effective batch size 32).
\noindent\textbf{Reward composition.}
$R = \lambda_1 R_{\mathrm{acc}} + \lambda_2 R_{\mathrm{fmt}} + \lambda_3 R_{\mathrm{rsn}}$
with $\lambda_1{:}\lambda_2{:}\lambda_3=0.6{:}0.2{:}0.2$
(\emph{accuracy-dominant}, motivated by PCSR's verifiable answers;
cf.~360-R1's balanced weighting). Ablation: Table~\ref{tab:RL_variants}, Variants~\ding{194}--\ding{197};
functional forms: Appendix~D.2.
\textbf{Output format.}
During RL training, we retain the same structured think--answer format used in evaluation, with reasoning enclosed in \texttt{<think>}...\texttt{</think>} and the final prediction in \texttt{<answer>}...\texttt{</answer>}.

\subsection{Group Relative Policy Optimization}
\label{sec:grpo}

Following the 360-R1 framework~\cite{zhang2025360rl}, we employ
Group Relative Policy Optimization
(GRPO)~\cite{shao2024deepseekmath,zhang2025360rl} as our reinforcement learning
objective. Compared with conventional Proximal Policy Optimization
(PPO)~\cite{schulman2017proximalpolicyoptimizationalgorithms}, which requires training an auxiliary
value network, GRPO derives its advantage signal directly from
group-level reward statistics, thereby reducing both computational
overhead and implementation complexity.

Concretely, given a question~$q$, the current policy
$\pi_{\mathrm{old}}$ produces a group of $G$~candidate
responses $\{o_1,\ldots,o_G\}$, each of which is assigned a
scalar reward $r_i$ by a reward model.
These raw scores are then standardized within the group
to yield zero-mean, unit-variance advantages:
\begin{equation}
\label{eq:reward_norm}
\hat{r}_i
  = \frac{r_i - \mathrm{mean}(\mathbf{r})}
         {\mathrm{std}(\mathbf{r})},
\end{equation}
which serve as a constant advantage estimate across all token
positions of the $i$-th response:
\begin{equation}
\label{eq:advantage}
\hat{A}_{i,t} = \hat{r}_i.
\end{equation}

Policy updates are carried out with a clipped surrogate objective to
bound the magnitude of each step:
\begin{equation}
\label{eq:clip_loss}
\mathcal{L}_{\mathrm{clip}}
  = -\mathbb{E}\Big[\min\big(
      \rho_t \,\hat{A}_t,\;
      \mathrm{clip}(\rho_t,\,1{-}\epsilon,\,1{+}\epsilon)\,
      \hat{A}_t
    \big)\Big],
\end{equation}
where the importance-sampling ratio is defined as
$\rho_t = \pi_\theta(o_t \mid q,\, o_{<t})
      \,/\,
      \pi_{\mathrm{old}}(o_t \mid q,\, o_{<t})$.
An additional KL penalty with respect to a reference policy
$\pi_{\mathrm{ref}}$ is imposed to regularize training and prevent
excessive policy drift:
\begin{equation}
\label{eq:grpo_loss}
\mathcal{L}_{\mathrm{GRPO}}(\theta)
  = \mathcal{L}_{\mathrm{clip}}
  + \beta \,\mathrm{KL}\!\left(
      \pi_\theta \,\|\, \pi_{\mathrm{ref}}
    \right),
\end{equation}
combining group-normalized advantages, clipped updates, and a KL 
penalty whose weight $\beta$ controls divergence strength against the reference policy $\pi_{\mathrm{ref}}$ (Eq.~\ref{eq:grpo_loss}), GRPO yields a lightweight optimization procedure well-suited to our rule-based reward setting. Detailed reward definitions and variant-specific weights: Appendix~D.

\begin{table}[t]
\centering
\caption{%
  RL variants and ablation settings under the unified \S5 stack. 
  ID indicates the variant family: V1-a/V1-b are matched non-RL controls; V2--V5 vary reward composition/weights; V6/V7 isolate new vs.\ old implementations
  of $R_{\mathrm{acc}}$/$R_{\mathrm{fmt}}$.
  Weights
  $\lambda_1{:}\lambda_2{:}\lambda_3$
  apply to
  $R_{\mathrm{acc}}{:}R_{\mathrm{fmt}}{:}R_{\mathrm{rsn}}$. 
  \textbf{Avg.} = mean over all eight tasks;
  H-Avg = mean (T2, T4, T5).
  Bold = best; \underline{underline} = second-best.
  All eight variants are reported to avoid selection bias.
  Variant~\ding{195}$^\star$ is the \textbf{H-Avg-optimal};
  Variant~\ding{196}$^\star$ is the
  \textbf{Avg-optimal}  (highest overall average).
}
\label{tab:RL_variants}
\scriptsize
\setlength{\tabcolsep}{0.5pt}
\renewcommand{\arraystretch}{1.08}

\begin{tabular}{@{}
>{\raggedright\arraybackslash}p{0.36cm}
>{\centering\arraybackslash}p{0.44cm}
>{\centering\arraybackslash}p{0.48cm}
>{\centering\arraybackslash}p{0.74cm}
>{\raggedright\arraybackslash}p{1.03cm}
>{\centering\arraybackslash}p{1.13cm}
>{\raggedright\arraybackslash}p{1.20cm}
!{\hspace{1pt}\color{gray!25}\vrule width 2pt\hspace{1pt}}
>{\centering\arraybackslash}p{0.49cm}
>{\centering\arraybackslash}p{0.49cm}
>{\centering\arraybackslash}p{0.49cm}
>{\centering\arraybackslash}p{0.49cm}
>{\centering\arraybackslash}p{0.62cm}
@{}}
\toprule
\multicolumn{7}{c}{\textbf{Variant definition}} &
\multicolumn{5}{c}{\textbf{Accuracy}} \\
\cmidrule(r){1-7}\cmidrule(l){8-12}

\textbf{ID}
& \textbf{$R_{\mathrm{acc}}$}
& \textbf{$R_{\mathrm{fmt}}$}
& \textbf{$R_{\mathrm{rsn}}$}
& \textbf{Setting}
& \makecell[c]{\textbf{$\lambda_1{:}\lambda_2{:}\lambda_3$}}
& \textbf{Purpose}
& \makecell[c]{\textbf{Avg.}}
& \textbf{T2}
& \textbf{T4}
& \textbf{T5}
& \textbf{H-Avg} \\
\midrule

\multicolumn{12}{@{}l}{\cellcolor{gray!12}\textit{Matched non-RL controls}} \\
\ding{192}-a
& --- & --- & ---
& \makecell[l]{Base-style\\ prompt}
& ---
& \makecell[l]{Matched\\ control}
& 31.28 & 13.71 & 6.73 & 21.22 & 13.89 \\

\ding{192}-b
& --- & --- & ---
& \makecell[l]{Train-style\\ prompt}
& ---
& \makecell[l]{Matched\\ control}
& 31.10 & 14.52 & 5.11 & 20.65 & 13.43 \\

\midrule
\multicolumn{12}{@{}l}{\cellcolor{gray!12}\textit{Reward / weight ablations}} \\
\ding{193}
& new & new & ---
& \makecell[l]{Reason-free\\RL}
& 0.50:0.50:0
& \makecell[l]{Weight\\ ablation}
& \underline{63.06} & 40.88 & 31.36 & \textbf{37.06} & 36.43 \\

\ding{194}
& new & new & new
& \makecell[l]{Balanced\\full reward}
& 0.40:0.40:0.20
& \makecell[l]{Weight\\ ablation}
& 34.80 & 15.09 & 27.59 & 27.07 & 23.25 \\

\rowcolor{yellow!12}
\ding{195}$^\star$
& new & new & new
& \makecell[l]{Main\\ full reward}
& 0.60:0.20:0.20
& \makecell[l]{Main variant\\ }
& 60.06 & \underline{47.20} & \textbf{42.80} & 30.87 & \textbf{40.29} \\

\ding{196}$^\star$
& new & new & ---
& \makecell[l]{Reason-free\\RL}
& 0.75:0.25:0
& \makecell[l]{Reason\\ ablation}
& \textbf{65.27} & \textbf{50.85} & 30.96 & \underline{35.02} & \underline{38.94} \\

\midrule
\multicolumn{12}{@{}l}{\cellcolor{gray!12}\textit{Implementation ablations}} \\
\ding{197}
& new & old & new
& \makecell[l]{Old $R_{\mathrm{fmt}}$\\ }
& 0.60:0.20:0.20
& \makecell[l]{Format\\ ablation}
& 29.56 & 14.19 & 24.09 & 20.65 & 19.64 \\

\ding{198}
& old & new & new
& \makecell[l]{Old $R_{\mathrm{acc}}$\\ }
& 0.60:0.20:0.20
& \makecell[l]{Accuracy\\ ablation}
& 41.34 & 15.57 & \underline{40.51} & 25.37 & 27.15 \\
\bottomrule
\end{tabular}
\end{table}

\noindent \textbf{Training Variants and Ablation Design}
To disentangle the effects of reward composition, weight allocation, and reward implementation, we organize the experiments into two matched non-RL controls and six RL variants (Table~\ref{tab:RL_variants}).
Variants~\ding{192}-a and~\ding{192}-b are matched non-RL controls under the same \S5 inference stack. 
Variants~\ding{193}--\ding{196} vary reward composition and weight allocation, while Variants~\ding{197} and~\ding{198} isolate the old/new implementations of $R_{\mathrm{fmt}}$ and $R_{\mathrm{acc}}$ under the same full-reward setting. This design separates matched controls from RL variants, since the goal of \S5 is not direct comparison with \S4, but to examine which reward choices most affect PCSR behavior under a controlled evaluation stack.
For space efficiency, Table~\ref{tab:RL_variants} reports only a compact cross-variant summary in the main text; Appendix~D contains the detailed reward definitions, metrics, and additional qualitative examples needed to avoid the impression of selective reporting.

\noindent\textbf{Training stability.}
Across 18 checkpoints, Variant~\ding{195} improves
from 37.0 to 60.93 despite intermediate fluctuations, 
with the
final three checkpoints (cp-1600/1700/1800 = 57.06/60.16/60.93)
forming a stable plateau.
Two
independent runs finish at 60.93 and 60.06 (0.87\,pp apart),
far below the 35.71\,pp spread across
Variants~\ding{193}--\ding{198}. This supports
reward-configuration sensitivity rather than run-to-run
instability (Appendix~D.4).

\subsection{Results and Analysis}
\label{sec:rl_results}

All comparisons below are \emph{within-stack} under the unified \S5 evaluation (Table~B.1); \S4 and \S5 numbers are not directly swappable.
This subsection consolidates our findings into three core conclusions. We demonstrate that the spatial reasoning deficits observed earlier are partially recoverable through RL, that this optimization improves internal logical consistency alongside aggregate scores, and that benchmark conclusions remain highly sensitive to the underlying evaluation protocol.

\noindent \textbf{Partial plasticity and the reward-design trade-off.}
First, we establish that the systematic failures identified in \S4 are at least partially plastic. Under a unified evaluation stack, the matched non-RL controls (Variants~\ding{192}-a and~\ding{192}-b) achieve similar relaxed accuracies (\(\sim\)31.2\%, Table~\ref{tab:RL_variants}), confirming that mere prompt rewriting cannot explain subsequent gains. Targeted RL substantially improves performance (up to 65.27\% for Variant~\ding{196}), proving that advanced PCSR behaviors can be recovered. However, this recovery is highly sensitive to both reward allocation and implementation. For instance, changing the reward weights (Variant~\ding{194} vs.\ \ding{195}) or replacing the structured format reward with a binary check (Variant~\ding{197}; see Appendix D for implementation details)
drastically swings overall accuracy from 29.56\% to 60.06\%.
Furthermore, the full-reward configuration (Variant~\ding{195}) reveals a critical task-dependent trade-off: while it does not maximize the global average, it achieves the highest performance (40.29\%) on the most demanding reasoning-heavy subset (T2/T4/T5). This indicates a \emph{targeted mechanism} that sacrifices some easy-task performance to strengthen core PCSR reasoning on T2/T4/T5 for one 7B MLLM; broad recoverability across scales and backbones remains open.

\begin{figure}[ht]
    \centering
    \includegraphics[width=0.9\columnwidth]{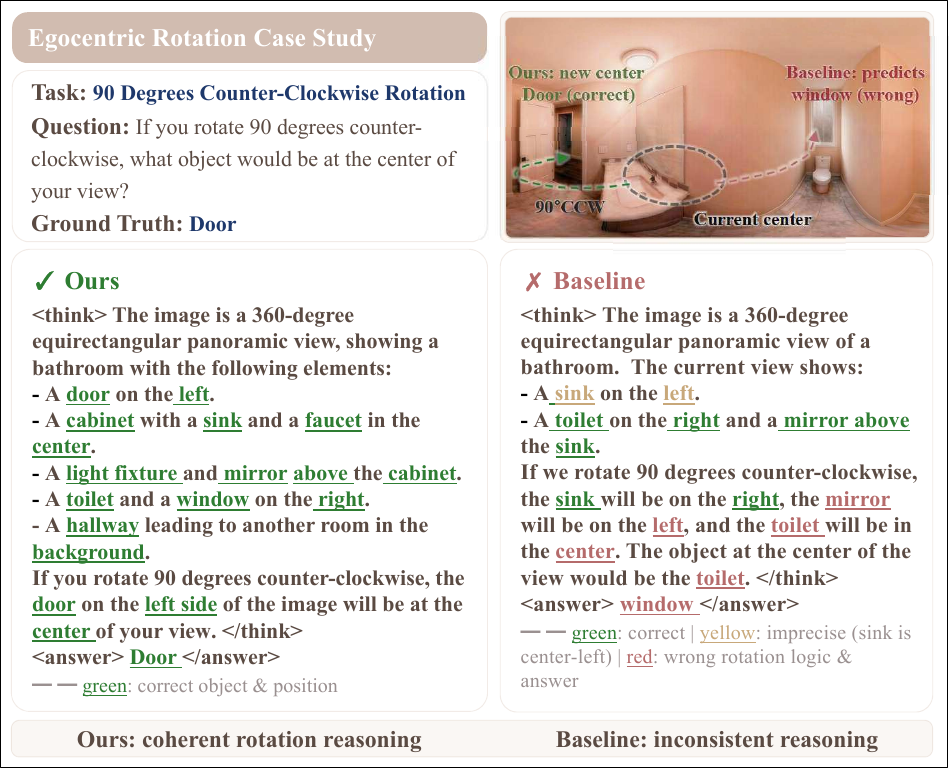}
    \caption{
    Qualitative comparison of reasoning processes in an egocentric rotation example. Given a 360$^\circ$ scene and the question of which object would move to the center after a 90$^\circ$ counter-clockwise rotation, our method correctly predicts \emph{door}, while the baseline predicts \emph{window}.The baseline also shows a reasoning--answer mismatch, whereas our method maintains a more consistent spatial update and a final answer that is consistent with its intermediate reasoning.
    }
    \Description{A qualitative comparison figure for an egocentric rotation task in a 360-degree panoramic bathroom scene. The figure contains a question asking which object would be at the center of the view after a 90-degree counter-clockwise rotation, together with the panoramic image. The left panel, labeled Ours, shows a reasoning trace that identifies the main bathroom objects and concludes that the door would move to the center; the final answer is Door and is marked correct. The right panel, labeled Baseline, shows a different reasoning trace with incorrect spatial update and inconsistency between the intermediate reasoning and the final answer; the final answer is window and is marked incorrect. A summary bar at the bottom highlights the contrast: Ours predicts Door correctly, while Baseline predicts window incorrectly.}
    \label{fig:qual_rotation_case}
\end{figure}

\noindent \textbf{Beyond task scores: RL improves reasoning--answer consistency.}
The benefits of reasoning-aware optimization extend beyond aggregate accuracy to the internal logical coherence of the model's generation. As illustrated in the egocentric rotation case study (Figure~\ref{fig:qual_rotation_case}), the baseline model not only predicts the wrong object but also exhibits a severe reasoning--answer mismatch: its intermediate text deduces that the \emph{toilet} would be at the center, yet its final output is \emph{window}. In contrast, our RL-optimized method maintains a coherent spatial update throughout its \texttt{<think>} trace and produces a final answer (\emph{door}) that perfectly aligns with its intermediate reasoning. This qualitative evidence complements the benchmark results, confirming that the reward mechanism successfully penalizes disjointed logic and enforces strict consistency between the spatial reasoning chain and the final prediction on complex 90\(^\circ\) viewpoint-conditioned transformations \cite{niu2026rotbench}.
This case-level observation is corroborated at scale by an
independent judge-based assessment of the full 9{,}697-item test
split (Appendix~D.3, Table~D.1
), which
rates the RL variant's reasoning traces 1.00 point higher on a 0--10
scale of image-direction understanding.

\noindent \textbf{Protocol sensitivity and methodological implications.}
Finally, the discrepancy between the earlier Qwen2.5-VL-7B zero-shot results (\S4) and the matched controls (\S5) highlights a crucial methodological takeaway: PCSR evaluation is highly protocol-sensitive. For tasks involving panoramic visual inputs and long-form structured outputs,
factors such as inference backend, visual preprocessing, generation budget, and answer parsing rules
materially shape the final benchmark conclusions;
detailed protocol differences are summarized in Appendix D.
Consequently, the numerical results from \S4 \&~5 cannot be rigidly aligned or directly swapped. Instead, these findings emphasize that the evaluation strategy is an intrinsic part of the benchmark itself. While the spatial reasoning limitations of MLLMs are partially recoverable, robust benchmarking requires explicit protocol documentation and strictly matched inference settings for valid cross-variant comparisons.

\section{Discussion}
\label{sec:discussion}
The findings of \S4–5 show that strong perceptual coverage does not guarantee robust spatial modeling \citep{yang2025thinking, wang2025mindcube}. While panoramic inputs remove field-of-view limits, they expose a deeper bottleneck: current MLLMs struggle to simulate spatial changes. Failures in viewpoint transformation, re-anchoring, and multi-step composition suggest that models treat spatial relations as static 2D correlations rather than dynamic 3D structures. 

Our RL analysis further reveals that this deficit is only partially plastic and comes with trade-offs. 
Strong reasoning rewards improve complex transformations but degrade basic
perception---a trade-off directly visible in our own ablation (Variant~\ding{194}
vs.~\ding{195}, Table~\ref{tab:RL_variants}), resembling a “reasoning tax” \cite{ren2026mitigatingreasoningtaxvisionlanguage}. The high sensitivity to reward design and the mismatch between intermediate reasoning and final outputs also highlight fragile spatial coherence, indicating reliance on shortcuts and the need for process-level supervision rather than outcome-based rewards.

Positioning our RL variants against supervised fine-tuning remains open in this pilot, but the current pattern is consistent with prior reports on reasoning-oriented multimodal post-training\cite{shao2024deepseekmath}: as reasoning depth increases,
reasoning-aware RL tends to outperform both LoRA and full-parameter SFT, with the gap widening on the hardest subsets.
Two internal observations anchor this on PCSR-Bench: the H-Avg-optimal variant in Table~\ref{tab:RL_variants} favors the hard subset (T2/T4/T5), whereas the Overall-optimal variant stays closer to the zero-shot baseline on the Perception group
(T0/T6/T7, Table~\ref{tab:zero_shot_relaxed}).
Together, these two variants indicate that hard-subset gains require a reasoning signal that plain next-token supervision cannot provide.

Under this lens, vanilla LoRA is expected to under-fit the perspective-transformation kernel of the hard subset 
and 
converge instead on surface-form regularities (answer-format compliance, in-vocabulary directional tokens); 
full-parameter fine-tuning faces the opposite risk of over-writing the perception ability that the zero-shot backbone already possesses
(Table~\ref{tab:zero_shot_relaxed}).
A matched-budget empirical comparison against LoRA and full-parameter SFT, carried out under the same evaluation stack, is a natural direction for future work.

The gap between earlier zero-shot results and the matched controls further shows that PCSR evaluation is highly protocol-sensitive: inference backend, visual preprocessing, and answer parsing each materially shift reported scores. The current benchmark scope is limited to indoor, single-image reasoning; outdoor panoramas and action-conditioned 360$^\circ$ video remain open extensions.

\section{Conclusion}
\label{sec:conclusion}
We introduced PCSR-Bench, a 360$^\circ$ benchmark to diagnose perspective-conditioned spatial reasoning in multimodal large language models from omnidirectional images. The benchmark reveals a clear gap: full-scene visual access improves coverage, but does not by itself yield robust viewpoint transformation, perspective re-anchoring, or compositional spatial reasoning.
Our RL-based diagnostic study further shows that this gap is partially plastic but not uniformly recoverable, yet the gains remain task-selective, reward-sensitive, and protocol-sensitive. Taken together, these findings do not overturn the benchmark diagnosis; rather, they show that PCSR is a genuine bottleneck with limited but non-negligible recoverability. Further progress likely requires stronger spatially grounded modeling, more effective reward design, and more carefully controlled evaluation protocols.

\begin{acks}
This study was supported by the Laboratory for Artificial Intelligence in Design and the InnoHK Initiative of the Innovation and Technology Commission of the Hong Kong Special Administrative Region Government, Project No. 2.4. The cloud computing resources and related expenses for this work were supported by the National Natural Science Foundation of China (No.~42571358). The authors also thank Ji Qi (Institute of Aerospace Remote Sensing Innovations and School of Geography and Remote Sensing, Guangzhou University, Guangzhou 510006, China) for proofreading the manuscript and providing helpful comments. During the preparation of this manuscript, the authors used ChatGPT and Gemini solely to improve the readability and flow of the language. After using these tools, the authors carefully reviewed and edited the text, and take full responsibility for the final content of the publication.
\end{acks}

\clearpage

\bibliographystyle{ACM-Reference-Format}
\balance
\bibliography{Chen7653}

\end{document}